\documentclass{article}

\usepackage[main,final]{neurips_2026}

\usepackage[utf8]{inputenc} %
\usepackage[T1]{fontenc}    %
\usepackage{hyperref}       %
\hypersetup{hidelinks}
\usepackage{url}            %
\usepackage{booktabs}       %
\usepackage{amsfonts}       %
\usepackage{nicefrac}       %
\usepackage{microtype}      %
\usepackage{xcolor}         %

\usepackage{multirow}
\usepackage{arydshln}
\usepackage{amsmath}
\usepackage{graphicx}
\usepackage{colortbl}
\usepackage{array}
\usepackage{makecell}
\newcounter{algorithm}
\renewcommand{\thealgorithm}{\arabic{algorithm}}

\newcommand{\upmark}[1][orange]{\rlap{\hspace{0.25em}\raisebox{.2ex}{\scriptsize\textcolor{#1}{$\uparrow$}}}}

\newcolumntype{L}[1]{>{\raggedright\arraybackslash}m{#1}}
\newcolumntype{C}[1]{>{\centering\arraybackslash}m{#1}}

\usepackage{amsthm}
\usepackage{float}
\newtheorem{definition}{Definition}[section]

\title{When Language Overrules: Revealing Text Dominance in Multimodal Large Language Models}

\author{%
  Huyu Wu$^{1}$ \quad Meng Tang$^{2}$ \quad Xinhan Zheng$^{1,3}$ \quad Duo Su$^{4}$ \quad Haiyun Jiang$^{5}$\thanks{Corresponding author.} \\
  {\small $^{1}$Beijing Angelpro Technology Co., Ltd. \quad $^{2}$Department of Computer Science, Aberystwyth University} \\
  {\small $^{3}$University of Science and Technology of China \quad $^{4}$Tsinghua University \quad $^{5}$Fudan University} \\
  \texttt{huyu-wu@outlook.com} \quad \texttt{haiyunjiangnlp@gmail.com}
}

\begin{document}

\maketitle

\begin{abstract}
Text dominance, the tendency of multimodal large language models (MLLMs) to over-attend to textual tokens while under-utilizing non-text inputs, has been observed in vision-language settings, but whether it extends to other modalities remains an open question.
We present a systematic, attention-level study of this phenomenon across five modalities (image, video, audio, time-series, and graph), using two diagnostic metrics: the Modality Dominance Index (MDI), which compares per-token attention between text and non-text inputs, and the Attention Efficiency Index (AEI), which normalizes attention share by token share.
Across ten model-benchmark pairs and five benchmarks, text dominance appears in the tested image, video, audio, and time-series settings and typically intensifies in deeper layers.
Controlled token-replication experiments show that expanding non-text sequences without adding semantic content amplifies the imbalance, while graph tasks provide a boundary case where compact, information-dense tokens reverse the effect.
Guided by these findings, we evaluate attention-based token compression as a proof-of-concept intervention on the vision modality.
On LLaVA-1.5-7B with MMMU-Pro, removing 90\% of visual tokens via informed selection preserves accuracy (33.70\% vs.\ 33.23\% baseline), while random dropping under the same ratio degrades it to 30.29\%; informed selection also lowers the late-layer MDI from 15.63 to 3.46 and reduces latency by 3.5$\times$.
These results suggest that text dominance is a cross-modal phenomenon associated with token redundancy, and that token compression can reduce it in the tested vision setting while preserving task performance.
\end{abstract}

\section{Introduction}
\label{sec:intro}

Multimodal large language models (MLLMs) \cite{yin2024survey, qin2025survey, team2025kimi, bai2025qwen2} have achieved strong results across vision, audio, and structured-data tasks by combining modality-specific encoders with a large language model backbone.
Despite these advances, a growing body of evidence suggests that the integration is far from balanced: MLLMs tend to allocate disproportionate attention to textual tokens while under-utilizing non-text inputs \cite{cai2025mllm, zheng2025mllms, jia2025symdpo}, a pattern we call \emph{text dominance}.

Evidence of this imbalance has so far been reported primarily in vision-language settings.
For instance, VQA models can answer questions correctly even when the input image is removed \cite{liu2024eliminating}, suggesting that text tokens alone carry sufficient signal for the model's prediction.
\cite{leng2024curse} proposed the Modality Importance Score for video question answering, extending the observation beyond static images.
However, these studies are limited to one or two modality pairs, leaving open the question of whether text dominance is specific to certain architectures or a more general property of current MLLMs.

\textit{In this work, we ask whether text dominance is merely an artifact of specific vision-language architectures, or a broader phenomenon that persists across modalities and network depths in current MLLMs.}

To answer this question, we conduct a systematic attention-level analysis across five modalities: image, video, audio, time-series, and graph.
We define two metrics: the Modality Dominance Index (MDI), which compares per-token attention between text and non-text inputs, and the Attention Efficiency Index (AEI), which normalizes each modality's attention share by its token share.

Our analysis yields three findings.
First, text dominance is pervasive: across ten models, it appears in image, video, audio, and time-series settings and intensifies in deeper layers, with the late-layer MDI reaching 157.53 for VideoLLaMA3-7B on MMBench-Video \cite{fang2024mmbench}.
Second, controlled token-replication experiments show that expanding non-text sequences without adding semantic content consistently amplifies the imbalance.
This supports the hypothesis that token redundancy drives attention dilution: as non-text tokens proliferate, per-token attention is spread thinner, favoring the more compact text modality.
Third, graph tasks provide a boundary case: when non-text tokens are compact and information-dense, the dominant modality can reverse, suggesting that the phenomenon depends on the interaction between token count and information density.

Motivated by these findings, we evaluate attention-guided token compression \cite{chen2024image, zhang2024cls} as a proof-of-concept intervention on the vision modality, where mature compression methods exist.
On LLaVA-1.5-7B with MMMU-Pro \cite{yue2024mmmu}, removing 90\% of visual tokens via informed selection preserves accuracy (33.70\% vs.\ 33.23\% baseline), while random dropping under the same ratio degrades it to 30.29\%.
The informed selection also reduces the late-layer MDI from 15.63 to 3.46, suggesting that many of the removed tokens were redundant in this setting.

To our knowledge, this is the first attention-level study of text dominance that spans five modalities and ten models under a unified diagnostic framework (MDI and AEI).
Beyond characterizing the phenomenon, we provide controlled evidence that non-text token proliferation amplifies text dominance, and observational evidence that task formulation further modulates the effect.
We also show, as a proof of concept on the vision modality, that existing token compression methods can reduce text dominance while preserving accuracy and lowering latency.

\begin{figure}[ht]
    \centering
    \includegraphics[width=0.4\textwidth]{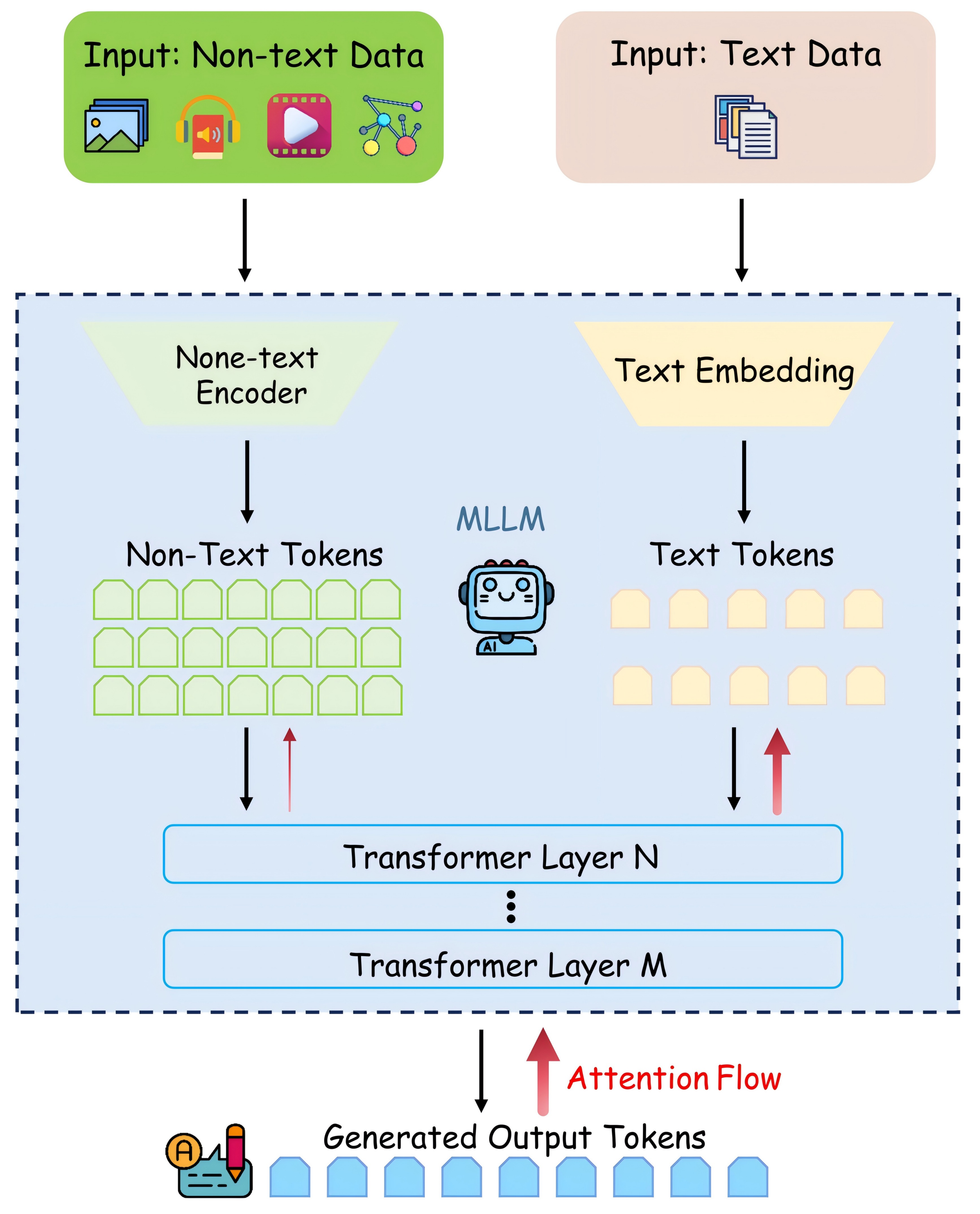}
    \caption{Overview of text dominance in MLLMs. Each modality is tokenized and fed to the LLM backbone. The thinner arrows from non-text tokens reflect their lower per-token attention: despite comprising the majority of input tokens, non-text modalities receive substantially less attention per token than text.}
    \label{fig1}
\end{figure}

\section{Related Work}
\label{sec:rw}

\subsection{Modality Imbalance: Evidence and Explanations}
MLLMs typically combine a pretrained modality-specific encoder with a large language model backbone via a learned interface \cite{liang2024survey}.
This architecture has been extended to video \cite{zhang2025videollama}, audio \cite{chu2024qwen2}, time-series \cite{xie2024chatts}, and graph data \cite{tang2024graphgpt}, yet a recurring observation across these settings is modality imbalance \cite{prabhu2025unveiling}: the model over-relies on text while under-utilizing non-text inputs.

The roots of this imbalance have been attributed to both data and architecture.
On the data side, the higher information density of text relative to raw perceptual input creates exploitable shortcuts \cite{park2025assessing}.
On the architecture side, most MLLMs pair an LLM pretrained on trillions of text tokens with a vision encoder trained on a comparatively smaller corpus \cite{li2023blip, liu2023visual}, creating an asymmetry in backbone capacity.
Prior work has studied this imbalance primarily in image-text settings, without attention-level quantification across modalities.
We provide such a cross-modal analysis, covering five modalities and ten model-benchmark pairs with a unified diagnostic framework.

\subsection{Token Efficiency and Compression in MLLMs}
A complementary line of work seeks to mitigate modality imbalance through training-time or inference-time interventions.
The Data Remixing framework \cite{ma2025improving} masks the stronger modality during training to force reliance on the weaker one.
The MBPO framework \cite{liu2025modality} applies Direct Preference Optimization on adversarially generated negatives to reduce text-driven hallucination.
At inference time, methods such as FastV \cite{chen2024image} and FasterVLM \cite{zhang2024cls} compress visual tokens to improve efficiency.
These approaches target efficiency or alignment but do not connect token compression to modality balance.
We repurpose existing compression methods as controlled tools to test the attention dilution hypothesis, rather than proposing a new compression algorithm.

\section{Text Dominance in Multimodal Large Language Models}
\label{sec:evaluation}

To move beyond qualitative observations of modality bias, we require a rigorous quantitative framework.
This section formalizes the measurement of text dominance: we first describe the benchmarks and models (\S\ref{subsec:datasets}), then define two diagnostic metrics (\S\ref{subsec:metrics}), and present the main results (\S\ref{subsec:results}).

\subsection{Datasets and Models}
\label{subsec:datasets}
We evaluate across five modalities, selecting benchmarks intended to reduce text-only shortcuts.
For image, we use MMMU-Pro \cite{yue2024mmmu}, which filters out text-only-answerable questions, and test Qwen2.5-VL (7B/32B/72B) \cite{bai2025qwen2}, LLaVA-1.5-7B \cite{liu2024improved}, and Kimi-VL-A3B-Instruct \cite{team2025kimi} to cover diverse model scales.
For video, MMBench-Video \cite{fang2024mmbench} provides long-form temporal reasoning tasks, evaluated on Qwen2.5-VL-7B and VideoLLaMA3-7B \cite{zhang2025videollama}.
Audio, time-series, and graph are each represented by one domain-standard benchmark, IEMOCAP \cite{busso2008iemocap}, TS-Reasoning \cite{xie2024chatts}, and GraphGPT-Eval \cite{tang2024graphgpt}, paired with the corresponding specialist model (Qwen2-Audio-7B-Instruct \cite{chu2024qwen2}, ChatTS-14B, and GraphGPT-7B).
Full details of all model-benchmark pairs are listed in Table~\ref{tab:mdi_aei}.

\subsection{Evaluation Metrics}
\label{subsec:metrics}

To characterize how an MLLM allocates its computational resources across modalities, we employ two complementary indices: the \textit{Modality Dominance Index} (MDI) and the \textit{Attention Efficiency Index} (AEI).
The MDI captures overall modality dominance in generation, whereas the AEI measures the attention efficiency of each modality relative to its token proportion.

Let $\mathcal{T}$ and $\mathcal{O}$ denote the sets of text tokens and non-text tokens, respectively.
For the generation of $N$ output tokens, let $a_{i,j}$ denote attention from output token \(i\) to input token \(j\), averaged over selected layers and all heads, then normalized over input tokens at each step.
We define the total attention mass received by text and non-text tokens as
\begin{equation}
A_T = \sum_{i=1}^{N} \sum_{j \in \mathcal{T}} a_{i,j},
\qquad
A_O = \sum_{i=1}^{N} \sum_{j \in \mathcal{O}} a_{i,j}.
\end{equation}

\noindent
\begin{definition}[Modality Dominance Index]
We define the Modality Dominance Index as
\begin{equation}
\mathrm{MDI}
=
\frac{A_T / |\mathcal{T}|}{A_O / |\mathcal{O}|}.
\label{eq:mdi}
\end{equation}
\end{definition}
The MDI compares the average attention received by one text token and one non-text token.
A value larger than $1$ indicates per-token text dominance, a value smaller than $1$ indicates per-token non-text dominance, and a value close to $1$ indicates similar average attention per token across the two modalities.

\noindent
\begin{definition}[Attention Efficiency Index]
Let
\begin{equation}
P_T = \frac{A_T}{A_T + A_O},
\qquad
Q_T = \frac{|\mathcal{T}|}{|\mathcal{T}| + |\mathcal{O}|},
\end{equation}
where $P_T$ is the attention share of text and $Q_T$ is the token share of text.
We define the Attention Efficiency Index for the text modality as
\begin{equation}
\mathrm{AEI}
=
\frac{P_T}{Q_T}
=
\frac{A_T / (A_T + A_O)}{|\mathcal{T}| / (|\mathcal{T}| + |\mathcal{O}|)}.
\label{eq:aei}
\end{equation}
\end{definition}

The AEI measures how effectively text converts its token budget into model attention.
An AEI value larger than $1$ indicates that text receives more attention than its token proportion would suggest, while a value smaller than $1$ indicates that text is under-attended relative to its input size.
Throughout this paper, AEI refers to the text modality unless otherwise specified.

MDI and AEI are complementary: MDI measures per-token dominance, while AEI normalizes attention by token budget; the two are related but not redundant.
Both are diagnostic statistics of attention allocation, not causal measures of modality importance \cite{jain2019attention}.
We therefore report them jointly with accuracy and latency throughout the paper, and interpret shifts only under controlled interventions (\S\ref{sec:analysis}) or matched baselines (\S\ref{sec:compression}).

\begin{table}[ht]
\centering
\footnotesize
\setlength\tabcolsep{3pt}
\begin{tabular}{ccc|cc|cc|cc}
\hline
\multirow{2}{*}{\textbf{Model}} & \multirow{2}{*}{\textbf{Modality}} & \multirow{2}{*}{\textbf{Dataset}} & \multicolumn{2}{|c|}{\textbf{Early}} & \multicolumn{2}{c|}{\textbf{Middle}} & \multicolumn{2}{c}{\textbf{Late}} \\
 &  &  & \textbf{MDI} & \textbf{AEI} & \textbf{MDI} & \textbf{AEI} & \textbf{MDI} & \textbf{AEI} \\
\hline
Qwen2.5-VL-7B        & \multirow{5}{*}{Image} & \multirow{5}{*}{MMMU\_Pro} & 2.26 & 1.42 & 21.12 & 10.86 & 33.10 & 14.24 \\
Qwen2.5-VL-32B        &                         && 3.84 &  2.82 &  54.96 & 21.88 &  26.03 & 13.95 \\
Qwen2.5-VL-72B        &                       &   & 9.33 &  6.15 & 92.21 & 60.43 &  24.46 & 14.60 \\
LLaVA-1.5-7B           &                         &  & 1.91 &  1.55 & 8.07 &  2.92 & 15.63 & 3.36 \\
Kimi-VL-A3B-Instruct      &                         &  & 2.27 &  3.91 &  3.78 &  2.99 & 28.39 & 2.59 \\
\cdashline{1-9}[1pt/5pt]
Qwen2.5-VL-7B        & \multirow{2}{*}{Video} &  \multirow{2}{*}{MMBench-Video} & 10.72 &  9.60 &  74.13 & 41.78 & 86.95 & 47.84 \\
VideoLLaMA3-7B            &                        &  & 19.14 & 17.90 & 140.10 & 73.75 & 157.53 & 76.26 \\
\cdashline{1-9}[1pt/5pt]
\multirow{3}{*}{Qwen2-Audio-7B-Instruct} & \multirow{3}{*}{Audio} & IEMOCAP $\times$1  & 1.02 & 1.32 &  3.24 & 1.99 &  1.16 & 1.08 \\
                                         &                        & IEMOCAP $\times$5  & 2.65 & 2.56 &  8.09 & 5.17 &  6.73 & 4.31 \\
                                         &                        & IEMOCAP $\times$10 & 2.80 & 2.50 & 10.10 & 5.46 &  8.70 & 5.09 \\
\cdashline{1-9}[1pt/5pt]
\multirow{3}{*}{ChatTS-14B} & \multirow{3}{*}{Time-series} & TS-Reasoning $\times$1  & 1.52 & 1.19 &  4.37 & 1.40 &  3.52 & 1.37 \\
                            &                              & TS-Reasoning $\times$5  & 2.08 & 1.95 & 10.72 & 3.15 &  9.28 & 3.03 \\
                            &                              & TS-Reasoning $\times$10 & 2.36 & 2.67 & 20.70 & 5.37 & 16.25 & 5.13 \\
\cdashline{1-9}[1pt/5pt]
\multirow{3}{*}{GraphGPT-7B} & \multirow{3}{*}{Graph} & GraphGPT-Eval $\times$1  & 0.14 & 0.84 & 0.14 & 0.84 & 0.20 & 0.90 \\
                             &                         & GraphGPT-Eval $\times$5  & 0.20 & 0.69 & 0.35 & 0.83 & 0.69 & 0.98 \\
                             &                         & GraphGPT-Eval $\times$10 & 0.31 & 0.71 & 0.68 & 0.97 & 1.35 & 1.14 \\
\hline
\end{tabular}
\caption{\textbf{MDI and AEI across models, modalities, and benchmarks.} Image and video rows use the original input without replication. The notation $\times n$ denotes the replication factor applied to non-textual tokens. Early, Middle, and Late report statistics from the first two, middle two, and last two model layers, respectively. GraphGPT-Eval and TS-Reasoning abbreviate GraphGPT-Eval-Instruction and TimeSeries-Reasoning. Rows with replication factors ($\times$5, $\times$10) are analyzed in \S\ref{sec:analysis}.}
\label{tab:mdi_aei}
\end{table}

\begin{figure}[htbp]
    \centering
    \includegraphics[width=0.4\textwidth]
    {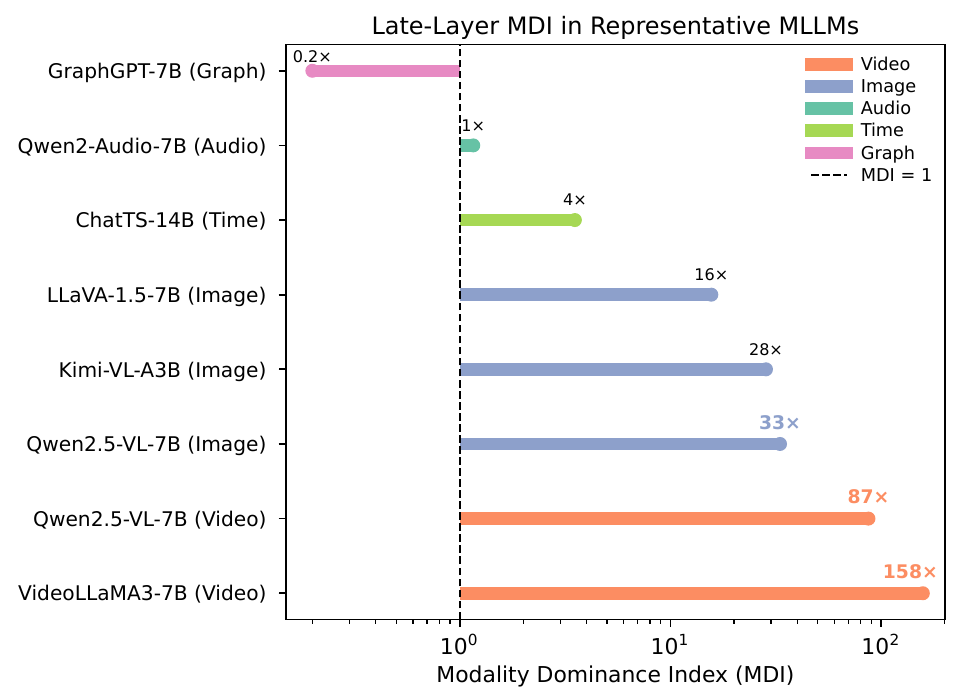}
    \caption{Text dominance across MLLMs of different modalities. The dashed line marks MDI\,=\,1; points to its right indicate text-dominant attention. The graph modality (MDI\,$<$\,1) is a boundary case discussed in \S\ref{subsec:graph}, where compact, information-dense tokens reverse the dominant modality.}
    \label{fig:Text dominance phenomenon}
\end{figure}

\subsection{Experimental Results}
\label{subsec:results}

Table~\ref{tab:mdi_aei} reports MDI and AEI across all model-benchmark pairs, aggregated over early, middle, and late layers; rows with replication factors ($\times$5, $\times$10) are analyzed in \S\ref{sec:analysis}.
The results reveal a clear and consistent pattern: text dominance is prevalent across most tested modalities, with graph as a notable exception, and typically intensifies in deeper layers.

\paragraph{Text dominance is pervasive in image and video tasks.}
For Qwen2.5-VL-7B on MMMU-Pro, the MDI rises from 2.26 in early layers to 33.10 in late layers, where image tokens receive 54.4\% of total input attention. VideoLLaMA3-7B reaches a late-layer MDI of 157.53 on MMBench-Video, with video tokens receiving 50.9\% of total input attention (Appendix~\ref{app:token_stats}).

We also observe that text dominance varies substantially across modalities and appears to covary with non-text token count (Table~\ref{tab:mdi_aei}); Section~\ref{sec:analysis} investigates the drivers of this variation through controlled experiments.

\begin{figure}[t]
    \centering
\includegraphics[width=0.68\textwidth]{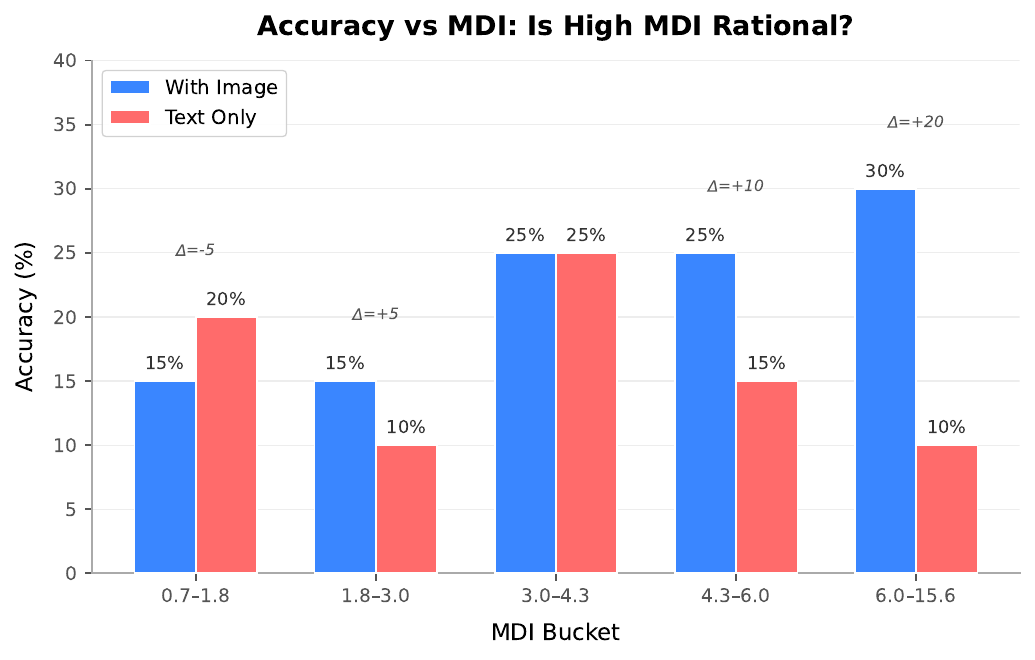}
    \caption{Accuracy under multimodal (\textit{With Image}) and text-only (\textit{Text Only}) inputs across MDI buckets on LLaVA-1.5-7B / MMMU-Pro. The accuracy gap $\Delta$ increases with MDI, indicating that the model under-attends to images most severely on samples where visual information is most critical for correct prediction.}
    \label{fig:mdi_bucket}
\end{figure}

\paragraph{Is text dominance aligned with task relevance?}
A natural question is whether text dominance reflects a rational allocation strategy, i.e., whether the model ignores non-text tokens precisely when they are uninformative.
To test this, we sort MMMU-Pro samples by their MDI on LLaVA-1.5-7B and divide them into five equal-sized buckets, then compare accuracy under the original multimodal input (\textit{With Image}) and a text-only ablation (\textit{Text Only}) that removes all image tokens (Figure~\ref{fig:mdi_bucket}).

The results reveal a striking paradox: the accuracy gap $\Delta$ between \textit{With Image} and \textit{Text Only} shows an overall increase from $-5$ percentage points in the lowest MDI bucket to $+20$ in the highest.
In other words, the samples where the model most heavily favors text tokens (MDI $\in [6.0, 15.6]$) are precisely the ones where images are most critical: removing images drops accuracy from 30\% to 10\%.
Conversely, in the lowest MDI bucket (MDI $\in [0.7, 1.8]$), images appear unnecessary or even harmful (accuracy improves from 15\% to 20\% without them), yet the model allocates relatively more attention to image tokens in this group.
This inverse relationship suggests that, at least in this setting, text dominance is not well-aligned with task relevance: the model under-attends to non-text tokens most severely when they carry the most decision-relevant information.
A qualitative case study illustrating this failure mode on individual samples is provided in Appendix~\ref{app:case_study}.

Having established that text dominance is pervasive and that attention allocation appears misaligned with task relevance, a natural question follows: what input factors are associated with stronger text dominance?
Section~\ref{sec:analysis} investigates this through controlled and observational evidence.

\section{What Drives Text Dominance? Empirical Evidence}
\label{sec:analysis}
The consistent escalation of MDI in deeper layers (Table~\ref{tab:mdi_aei}) suggests that input characteristics shape text dominance during generation.
We hypothesize that a major contributor is \emph{attention dilution} stemming from token redundancy: non-text modalities are encoded into far more tokens than the accompanying text, yet many of these tokens are semantically redundant, causing attention to spread thinly across them.
This section examines two factors that are consistently associated with stronger text dominance in our data: (i) token redundancy in non-text modalities, tested through controlled replication experiments, and (ii) text-centric task formulations, examined through observational comparison.
Graph tasks provide a boundary case that further supports the token-redundancy account (\S\ref{subsec:graph}).
For the first factor, we repeat the same non-text representations while varying token count, measuring the combined effect of sequence expansion and repeated content in the tested settings.
For the second factor, we present observational evidence that is suggestive but not causally identified, since task formulation co-varies with other design choices.

\subsection{Token Redundancy and Attention Dilution}
Non-text modalities are typically encoded into far more tokens than the accompanying text prompt, yet many of these tokens carry overlapping information.
If redundancy dilutes per-token attention, then expanding the non-text sequence without adding information should amplify text dominance.

We test this prediction with controlled token replication: we fix the text prompt and repeat the non-text token sequence by factors of $\times$1, $\times$5, and $\times$10 (Table~\ref{tab:mdi_aei}).
Across three modalities, late-layer MDI rises monotonically: Qwen2-Audio-7B-Instruct (1.16 $\to$ 6.73 $\to$ 8.70), ChatTS-14B (3.52 $\to$ 9.28 $\to$ 16.25), and GraphGPT-7B (0.20 $\to$ 0.69 $\to$ 1.35).
The last case is informative: a model that is initially non-text-dominant (MDI\,=\,0.20) becomes text-dominant (MDI\,=\,1.35) solely through token expansion, without any change in semantic content.

The video setting provides the most extreme natural illustration of this pattern.
VideoLLaMA3-7B encodes hundreds of frame tokens per video, reaching a late-layer MDI of 157.53 on MMBench-Video, with a corresponding AEI of 76.26.
Each text token receives substantially more attention on average, while video tokens collectively receive 50.9\% of total input attention in the last two layers.

\subsection{Task Formulation as an Additional Factor}
Task formulation appears to be an additional factor associated with text dominance beyond token count alone.
In time-series tasks, key normalization factors and task-specific metadata are encoded in natural language instructions, making the text prompt structurally central to inference.
Audio emotion recognition tasks similarly define the reasoning scaffold through textual prompts.

\begin{figure}[t]
    \centering
    \includegraphics[width=0.95\textwidth]{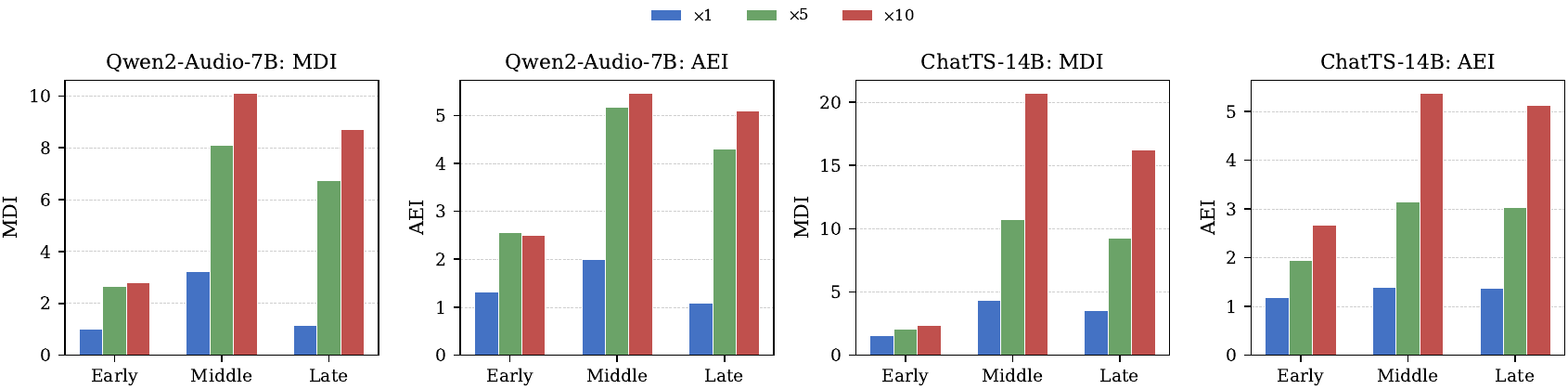}
    \caption{MDI and AEI of audio and time-series models under $\times$1, $\times$5, and $\times$10 token replication. Even at $\times$1, both models show AEI\,$>$\,1 in late layers, suggesting that task formulation contributes to text dominance beyond token scaling alone.}
    \label{fig:mdi_aei_a_ts}
\end{figure}

Even at $\times$1 replication (Figure~\ref{fig:mdi_aei_a_ts}), the textual modality already achieves AEI\,$>$\,1 in late layers for both ChatTS-14B (1.37) and Qwen2-Audio-7B-Instruct (1.08).
As non-text tokens are replicated, this advantage becomes more pronounced: for ChatTS-14B, the late-layer AEI increases from 1.37 ($\times$1) to 3.03 ($\times$5) and 5.13 ($\times$10).
A similar trend holds for Qwen2-Audio-7B-Instruct, where the late-layer AEI rises from 1.08 to 4.31 and 5.09 under the same conditions.

We note that this observation is correlational: we cannot fully separate the influence of task formulation from other factors such as pretraining data composition.
The key takeaway is that text dominance can be present even when non-text tokens are not heavily replicated, which suggests that token redundancy alone does not fully account for the phenomenon.

\subsection{Graph Tasks: A Boundary Case}
\label{subsec:graph}
Our framework predicts that if non-text tokens are compact and information-dense, the dominance should reverse.
Graph tasks are consistent with this prediction.
Under standard conditions, GraphGPT-7B yields a late-layer MDI of 0.20, with textual AEI at 0.90, indicating that graph tokens attract more attention per token than text.
When we inflate the graph token count by $\times$5 and $\times$10 replication without altering semantic content, MDI rises to 0.69 and 1.35, respectively, crossing the equilibrium threshold.
The textual AEI shifts correspondingly from 0.90 to 0.98 and 1.14.

These observations collectively suggest that token count and information density are important determinants of modality dominance in the tested settings, consistent with the attention dilution hypothesis.
Together with the token-scaling results above, these findings provide converging evidence that text dominance is associated with the expansion of semantically redundant non-text tokens.
This observation naturally motivates our intervention strategy in the next section.

\section{Can Token Compression Mitigate Text Dominance?}
\label{sec:compression}
The attention dilution hypothesis (\S\ref{sec:analysis}) predicts that reducing redundant non-text tokens should rebalance MDI.
This section tests that prediction on the vision modality, where mature compression methods exist and where text dominance is both strong and well-characterized in our diagnostic results.
We use existing methods as controlled tools to validate the attention dilution hypothesis and to assess the functional consequences of restoring attention balance (accuracy, latency).

\subsection{Experimental Setup}
\label{subsec:compression_setup}
We compare three token-reduction strategies under matched compression ratios (50\%, 75\%, 90\%) on LLaVA-1.5 (7B and 13B) with MMMU-Pro:
\begin{itemize}
\item \textbf{Random dropping}: uniformly samples a subset of visual tokens, serving as a \emph{mechanical baseline} that reduces token count without using any informativeness signal.
\item \textbf{FastV} \cite{chen2024image}: prunes visual tokens at an intermediate LLM layer based on the attention they receive during the first few decoding steps.
\item \textbf{FasterVLM} \cite{zhang2024cls}: ranks visual tokens by the attention they receive from the \texttt{[CLS]} token in the final layer of the frozen vision encoder, and retains the top-ranked tokens before fusion with text embeddings.
\end{itemize}
This comparison separates two effects: matched compression ratios test whether reducing token count alone lowers MDI, while accuracy differences between random and informed selection indicate that \emph{which} tokens are retained also matters beyond token count.
Formal algorithmic details of the attention-guided selection procedure are provided in Appendix~\ref{app:compression_alg}.

\subsection{Results}
\label{subsec:compression_results}

\begin{table}[t]
\centering
\footnotesize
\setlength\tabcolsep{5pt}
\renewcommand{\arraystretch}{1.05}
\begin{tabular}{l c l c c c c c c c c}
\toprule
\multirow{2}{*}{\textbf{Model}} & \multirow{2}{*}{\textbf{Reduction}} & \multirow{2}{*}{\textbf{Method}} & \multicolumn{2}{c}{\textbf{Early}} & \multicolumn{2}{c}{\textbf{Middle}} & \multicolumn{2}{c}{\textbf{Late}} & \multirow{2}{*}{\textbf{Acc}} & \multirow{2}{*}{\textbf{Time}} \\
\cmidrule(lr){4-5}\cmidrule(lr){6-7}\cmidrule(lr){8-9}
 &  &  & \textbf{MDI} & \textbf{AEI} & \textbf{MDI} & \textbf{AEI} & \textbf{MDI} & \textbf{AEI} &  &  \\
\midrule

\multirow{10}{*}{LLaVA-7B}
& 0\,\%  & No reduction & 1.91 & 1.55 & 8.07 & 2.92 & 15.63 & 3.36 & 33.23 & 0.42 \\
\cmidrule(lr){2-11}
& \multirow{3}{*}{50\,\%} & Random    & 1.38 & 1.30 & 4.14 & 1.84 & 8.98 & 2.28 & 31.30 & 0.15 \\
&                         & FastV & 1.28 & 1.23 & 4.11 & 1.84 & 8.91 & 2.23 & 33.23 & 0.17 \\
&                         & FasterVLM & 1.21 & 1.11 & 4.08 & 1.83 & 8.84 & 2.14 & 33.30\upmark & 0.15 \\
\cmidrule(lr){2-11}
& \multirow{3}{*}{75\,\%} & Random    & 0.96 & 1.03 & 2.38 & 1.41 & 5.74 & 1.64 & 31.79 & 0.14 \\
&                         & FastV & 0.95 & 1.01 & 2.41 & 1.44 & 5.78 & 1.66 & 33.17 & 0.16 \\
&                         & FasterVLM & 0.92 & 0.97 & 2.29 & 1.32 & 5.77 & 1.55 & 33.58\upmark & 0.14 \\
\cmidrule(lr){2-11}
& \multirow{3}{*}{90\,\%} & Random    & 0.82 & 0.87 & 1.38 & 1.15 & 3.49 & 1.29 & 30.29 & 0.12 \\
&                         & FastV & 0.79 & 0.96 & 1.31 & 1.09 & 3.51 & 1.32 & 33.35 & 0.14 \\
&                         & FasterVLM & 0.77 & 0.94 & 1.28 & 1.05 & 3.46 & 1.20 & 33.70\upmark & 0.12 \\

\midrule

\multirow{10}{*}{LLaVA-13B}
& 0\,\%  & No reduction & 5.67 & 2.64 & 16.92 & 3.52 & 17.68 & 3.56 & 34.62 & 0.27 \\
\cmidrule(lr){2-11}
& \multirow{3}{*}{50\,\%} & Random    & 3.48 & 1.92 & 8.28 & 2.19 & 9.35 & 2.27 & 32.71 & 0.21 \\
&                         & FastV & 3.25 & 1.80 & 8.14 & 2.15 & 9.24 & 2.20 & 34.52 & 0.22 \\
&                         & FasterVLM & 3.09 & 1.68 & 8.02 & 2.11 & 9.13 & 2.15 & 34.86\upmark & 0.21 \\
\cmidrule(lr){2-11}
& \multirow{3}{*}{75\,\%} & Random    & 2.34 & 1.39 & 4.72 & 1.59 & 6.20 & 1.66 & 32.15 & 0.19 \\
&                         & FastV & 2.22 & 1.34 & 4.58 & 1.54 & 6.14 & 1.60 & 33.85 & 0.20 \\
&                         & FasterVLM & 2.14 & 1.30 & 4.50 & 1.50 & 6.08 & 1.56 & 34.05\upmark & 0.19 \\
\cmidrule(lr){2-11}
& \multirow{3}{*}{90\,\%} & Random    & 1.73 & 1.05 & 2.54 & 1.23 & 3.78 & 1.28 & 31.52 & 0.18 \\
&                         & FastV & 1.65 & 1.12 & 2.43 & 1.18 & 3.72 & 1.23 & 33.90 & 0.19 \\
&                         & FasterVLM & 1.60 & 1.10 & 2.37 & 1.15 & 3.68 & 1.20 & 34.28\upmark & 0.18 \\

\bottomrule
\end{tabular}
\vspace{4pt}

\caption{\textbf{Effect of token-reduction methods and ratios on MDI, AEI, accuracy, and latency} (LLaVA-1.5 on MMMU-Pro). 0\% denotes the original setting. Under each non-zero ratio, we compare Random, FastV, and FasterVLM. {\scriptsize\textcolor{orange}{$\uparrow$}} indicates FasterVLM outperforms both Random and FastV in accuracy.}

\label{tab:reduction_compare_mdi_aei_acc_time}
\end{table}

\paragraph{Token compression rebalances MDI.}
As shown in Table~\ref{tab:reduction_compare_mdi_aei_acc_time}, all three strategies reduce MDI monotonically with the compression ratio, supporting the view that token count is a major driver in this setting.
At 90\% compression on LLaVA-7B, the late-layer MDI drops from 15.63 (baseline) to 3.49 (random), 3.51 (FastV), and 3.46 (FasterVLM).
The AEI exhibits a similar decreasing trend, indicating that the disproportionate attention advantage of text tokens diminishes as redundant visual tokens are removed.

\paragraph{Informed selection preserves accuracy better than random dropping.}
The gap between informed and random selection shows that token quality matters beyond token count.
On LLaVA-7B, random dropping at 90\% compression degrades accuracy to 30.29\%, while FastV maintains 33.35\% and FasterVLM achieves 33.70\% (baseline: 33.23\%).

\begin{figure}[t]
    \centering
\includegraphics[width=0.95\textwidth]{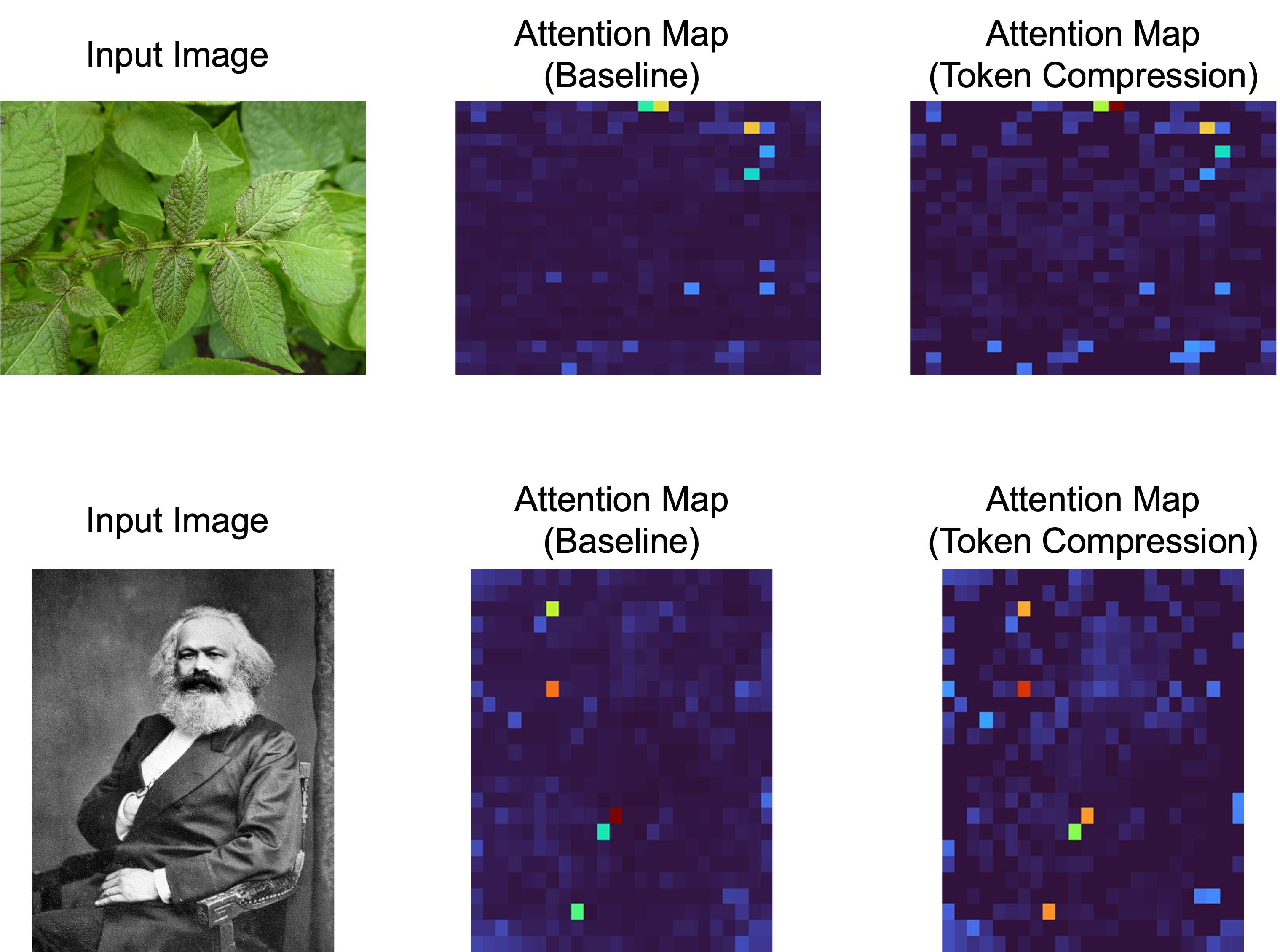}%
    \caption{Attention maps under 50\% token compression on two representative examples.
    From left to right: input image, baseline attention map, and attention map with 50\% compression.
    In both cases, the baseline prediction is incorrect, while token compression yields the correct answer.}
    \label{fig:compression_attn_maps}
\end{figure}

Visual inspection of attention maps (Figure~\ref{fig:compression_attn_maps}) illustrates that retained tokens correspond to semantically relevant regions.

Among the informed methods, the relative ranking varies with model size and compression ratio.
FasterVLM achieves the highest accuracy across all tested ratios on both LLaVA-7B and LLaVA-13B, although the margins over FastV are small (e.g., 34.05 vs.\ 33.85 at 75\% on LLaVA-13B) and should be interpreted cautiously.
Compression also reduces latency substantially, from 0.42\,s to 0.12\,s on LLaVA-7B and from 0.27\,s to 0.18\,s on LLaVA-13B at 90\% compression, while informed selection achieves comparable speed to random dropping at matched token budgets.
Overall, in this vision setting, token compression substantially lowers MDI, and informed selection preserves accuracy better than random dropping while retaining comparable efficiency gains.

\section{Discussion}
\label{sec:discussion}

\paragraph{Synthesizing the evidence.}
Our results across five modalities suggest that text dominance is a recurring pattern associated with the interaction between token count and information density.
When non-text tokens are numerous and redundant (image, video, audio, time-series), per-token attention is diluted and text dominance emerges; graph tasks provide a boundary case where compact, information-dense non-text tokens can reverse the effect.
In the image setting, the MDI bucket analysis (\S\ref{subsec:results}) further suggests a misalignment between attention allocation and task relevance: on LLaVA-1.5-7B with MMMU-Pro, the model tends to under-attend to image tokens more strongly on samples where removing images causes the largest accuracy drop.
While this evidence is drawn from a single model--benchmark pair and should be interpreted with caution, it is consistent with the broader pattern observed across modalities.

\paragraph{Implications for MLLM design.}
The compression experiments in \S\ref{sec:compression} show that, at least in the vision setting, removing up to 90\% of visual tokens can lower MDI while maintaining accuracy.
This suggests that the visual encoders tested here may produce more tokens than necessary for downstream reasoning, and that \emph{token efficiency in non-text encoders may deserve attention alongside model scaling} as a design consideration.
Whether this observation extends to other modalities and encoder families remains to be verified.

\paragraph{Scope and limitations.}
Several limitations should guide interpretation.
MDI and AEI are diagnostic statistics of attention allocation, not causal measures of modality importance; we mitigate this by jointly reporting accuracy throughout the paper.
Our token-replication experiments hold semantic content approximately fixed, but exact invariance cannot be guaranteed, as repeated tokens may affect layer normalization or positional encoding statistics.
Cross-model comparisons (e.g., LLaVA vs.\ Qwen2.5-VL) are observational, since architecture, training data, and optimization co-vary.

\section{Conclusion}
We conducted a systematic, attention-level study of text dominance across five modalities in MLLMs, using two diagnostic metrics (MDI and AEI).
Text dominance is pervasive across image, video, audio, and time-series settings, intensifies with non-text token count, and can reverse in graph tasks where non-text tokens are compact.
On the vision modality, attention-guided token compression reduces MDI while preserving accuracy and lowering latency, supporting the connection between token redundancy and attention imbalance.
Validating these findings on additional modalities, model families, and benchmarks is an important next step.

\renewcommand{\acksection}{\section*{Acknowledgments}}
\begin{ack}
This work was supported in part by the NSFC Special Fund and Major Program under Grants 62550004 and 62595773; in part by the Fundamental and Interdisciplinary Disciplines Breakthrough Plan of the Ministry of Education of China (No. JYB2025XDXM101); in part by the CPSF (No. 2025M771546).
\end{ack}

{
\small
\bibliographystyle{plainnat}
\bibliography{ref}
}

\appendix

\renewcommand{\thefigure}{S\arabic{figure}}
\setcounter{figure}{0}

\clearpage
\section{Experimental Environment}
\label{app:env}
The system was equipped with an Intel(R) Xeon(R) Gold 6348 CPU operating at a base frequency of 2.60\,GHz.
For accelerating deep learning computations, an NVIDIA A800 Tensor Core GPU with 80\,GB of dedicated memory was utilized.
All models were evaluated using Python 3.12.3 and PyTorch 2.3.0 with CUDA 12.1 support.

\section{Data Preprocessing Parameters}
\label{app:preprocess}
To manage hardware limitations, data was preprocessed as follows.
Images were dynamically resized to a total pixel count between $256 \times 28 \times 28$ and $1280 \times 28 \times 28$.
For videos, processing was model-specific: VideoLLaMA3-7B frames were sampled at 1\,fps up to a maximum of 90 frames and resized to $112 \times 112$; Qwen2.5-VL-7B frames were also sampled at 1\,fps but resized to a maximum of $100 \times 140$ pixels.

\section{Modality Attention Calculation}
\label{app:attention_calc}
To facilitate the calculation of the Modality Dominance Index (MDI) and the Attention Efficiency Index (AEI), we use the procedure detailed in Algorithm~\ref{alg:modality_attention} to acquire the per-token attention scores for text and non-text inputs.
This algorithm aggregates the attention weights from a specific output token to all input tokens, across designated layers and all their attention heads.
After normalization and partitioning by modality, the process yields a text attention score $S_{\text{text}}$ and a non-text attention score $S_{\text{non\_text}}$ for that single output token.
We first average the attention statistics over the generated tokens of each sample, then across samples. MDI is the ratio of the resulting mean per-token attention scores. AEI is computed from the mean attention masses and mean input token counts. For the Early, Middle, and Late statistics in Table~\ref{tab:mdi_aei}, we use the first two, middle two, and last two model layers, respectively.

\begin{figure}[ht]
\refstepcounter{algorithm}
\centering
\begin{minipage}{0.95\linewidth}
\rule{\linewidth}{0.8pt}\\[2pt]
{\small\textbf{Algorithm \thealgorithm} Modality Attention Calculation for an Output Token.}\label{alg:modality_attention}\\[2pt]
\rule{\linewidth}{0.4pt}\\[4pt]
{\small\textbf{Require:} Model $M$ (with $H$ heads); layer indices $L$; output token $o$; input text tokens $\mathcal{T}$; input non-text tokens $\mathcal{O}$.}\\[2pt]
{\small\textbf{Output:} Text attention score $S_{\text{text}}$; non-text score $S_{\text{non\_text}}$.}\\[4pt]
{\small
\hspace{0.5em}1:\quad Let $A^{(k,h)}_i$ be the attention from $o$ to input token $i$ at layer $k \in L$, head $h$.\\[2pt]
\hspace{0.5em}2:\quad Compute mean attention: $\bar{a}_i = \frac{1}{|L| \cdot H} \sum_{k \in L} \sum_{h=1}^{H} A^{(k,h)}_i$.\\[2pt]
\hspace{0.5em}3:\quad Normalize: $a_{\text{norm}} = \bar{a} \,/\, \|\bar{a}\|_1$.\\[2pt]
\hspace{0.5em}4:\quad $S_{\text{text}} = \frac{1}{|\mathcal{T}|} \sum_{i \in \text{indices}(\mathcal{T})} (a_{\text{norm}})_i$.\\[2pt]
\hspace{0.5em}5:\quad $S_{\text{non\_text}} = \frac{1}{|\mathcal{O}|} \sum_{i \in \text{indices}(\mathcal{O})} (a_{\text{norm}})_i$.\\[2pt]
\hspace{0.5em}6:\quad \textbf{Return} $S_{\text{text}},\; S_{\text{non\_text}}$.}\\[4pt]
\rule{\linewidth}{0.8pt}
\end{minipage}
\end{figure}

\section{Attention-Guided Visual Token Compression: Algorithmic Details}
\label{app:compression_alg}
This appendix provides the formal procedure for the attention-guided token compression used in \S\ref{sec:compression} (FasterVLM \cite{zhang2024cls}).

Let the frozen vision encoder output a sequence of patch tokens $V=\{v_1,v_2,\dots,v_N\}$ together with a class token $[\mathrm{CLS}]$.
For each patch token $v_i$, the saliency score is the normalized attention it receives from the class token in the final encoder layer:
\begin{equation}
s_i = \frac{1}{H} \sum_{h=1}^{H} \mathrm{Attn}^{(h)}_{[\mathrm{CLS}] \rightarrow i}, \qquad
\tilde{s}_i = \frac{s_i}{\sum_{j=1}^{N} s_j},
\end{equation}
where $H$ is the number of attention heads.
Given a compression ratio $r \in [0,1)$, we retain the top $M = \lfloor N(1 - r) \rfloor$ tokens by saliency, forming the compressed set $V' = \{\,v_i \mid \tilde{s}_i \ge \tilde{s}_{(M)}\,\}$, and discard the rest before fusion with textual embeddings.
Algorithm~\ref{alg:compression} summarizes the procedure.

\begin{figure}[ht]
\refstepcounter{algorithm}
\centering
\begin{minipage}{0.95\linewidth}
\rule{\linewidth}{0.8pt}\\[2pt]
{\small\textbf{Algorithm \thealgorithm} Attention-Guided Visual Token Compression \cite{zhang2024cls}.}\label{alg:compression}\\[2pt]
\rule{\linewidth}{0.4pt}\\[4pt]
{\small\textbf{Require:} Vision encoder output $V = \{v_1, \ldots, v_N\}$; class token $[\mathrm{CLS}]$; compression ratio $r$.}\\[4pt]
{\small
\hspace{0.5em}1:\quad Compute per-head attention: $a_i^{(h)} = \mathrm{Attn}^{(h)}_{[\mathrm{CLS}] \to i}$.\\[2pt]
\hspace{0.5em}2:\quad Average across heads: $s_i = \frac{1}{H}\sum_{h=1}^{H} a_i^{(h)}$.\\[2pt]
\hspace{0.5em}3:\quad Normalize: $\tilde{s}_i = s_i \,/\, \sum_j s_j$.\\[2pt]
\hspace{0.5em}4:\quad Compute budget: $M = \lfloor N(1-r) \rfloor$.\\[2pt]
\hspace{0.5em}5:\quad Retain top-$M$ tokens by $\tilde{s}_i$: $V' = \mathrm{TopK}(V, M)$.\\[2pt]
\hspace{0.5em}6:\quad \textbf{Return} compressed visual set $V'$.}\\[4pt]
\rule{\linewidth}{0.8pt}
\end{minipage}
\end{figure}

\section{Model Attention and Token Statistics}
\label{app:token_stats}
This section presents a detailed analysis of attention distribution and token count metrics across various models and data modalities.
To understand how these characteristics evolve through the model's depth, we segment our analysis by examining data from three distinct stages: the first two layers (Table~\ref{tab:consolidated_stats_early}), the middle two layers (Table~\ref{tab:consolidated_stats_middle}), and the last two layers (Table~\ref{tab:consolidated_stats_last}).
The tables summarize total attention and average attention per token for textual and non-textual inputs across the three layer groups.

\begin{table}[ht]
\centering
\footnotesize
\setlength\tabcolsep{3pt}
\caption{Attention and token statistics for the \textit{first two layers}.}
\label{tab:consolidated_stats_early}
\begin{tabular}{ll|cc|cc|ccc}
\toprule
\multirow{2}{*}{\textbf{Modality}} & \multirow{2}{*}{\textbf{Model}} & \multicolumn{2}{c|}{\textbf{Total Attn}} & \multicolumn{2}{c|}{\textbf{Avg Attn/Tok}} & \multicolumn{3}{c}{\textbf{Avg Token Count}} \\
\cmidrule(lr){3-4} \cmidrule(lr){5-6} \cmidrule(lr){7-9}
 & & \textbf{Text} & \textbf{Non-t} & \textbf{Text} & \textbf{Non-t} & \textbf{Text} & \textbf{Non-t} & \textbf{Out} \\
\midrule
\multirow{3}{*}{Image} & Qwen2.5-VL-7B & 0.049 & 0.951 & 6.0e-4 & 2.7e-4 & 118.6 & 3584.0 & 208.0 \\
 & Kimi-VL-A3B & 0.715 & 0.285 & 1.1e-2 & 1.3e-3 & 92.6 & 413.6 & 138.7 \\
 & LLaVA-1.5-7B & 0.175 & 0.825 & 2.3e-3 & 1.4e-3 & 117.3 & 576.0 & 67.7 \\
\cdashline{1-9}[1pt/5pt]
\multirow{2}{*}{Video} & Qwen2.5-VL-7B & 0.092 & 0.908 & 2.7e-3 & 2.5e-4 & 34.8 & 3584.0 & 77.6 \\
 & VideoLLaMA3-7B & 0.109 & 0.891 & 8.6e-3 & 4.5e-4 & 14.1 & 2288.4 & 15.8 \\
\cdashline{1-9}[1pt/5pt]
Audio & Qwen2-Audio-7B & 0.480 & 0.520 & 6.8e-3 & 6.7e-3 & 70.9 & 123.4 & 7.1 \\
\cdashline{1-9}[1pt/5pt]
Time Series & ChatTS-14B & 0.802 & 0.198 & 1.2e-2 & 7.9e-3 & 67.1 & 32.5 & 198.1 \\
\cdashline{1-9}[1pt/5pt]
Graph & GraphGPT-7B & 0.811 & 0.189 & 1.5e-3 & 1.1e-2 & 596.9 & 24.5 & 11.2 \\
\bottomrule
\end{tabular}
\end{table}

\begin{table}[ht]
\centering
\footnotesize
\setlength\tabcolsep{3pt}
\caption{Attention and token statistics for the \textit{middle two layers}.}
\label{tab:consolidated_stats_middle}
\begin{tabular}{ll|cc|cc|ccc}
\toprule
\multirow{2}{*}{\textbf{Modality}} & \multirow{2}{*}{\textbf{Model}} & \multicolumn{2}{c|}{\textbf{Total Attn}} & \multicolumn{2}{c|}{\textbf{Avg Attn/Tok}} & \multicolumn{3}{c}{\textbf{Avg Token Count}} \\
\cmidrule(lr){3-4} \cmidrule(lr){5-6} \cmidrule(lr){7-9}
 & & \textbf{Text} & \textbf{Non-t} & \textbf{Text} & \textbf{Non-t} & \textbf{Text} & \textbf{Non-t} & \textbf{Out} \\
\midrule
\multirow{3}{*}{Image} & Qwen2.5-VL-7B & 0.348 & 0.652 & 3.8e-3 & 1.8e-4 & 118.6 & 3584.0 & 208.0 \\
 & Kimi-VL-A3B & 0.547 & 0.453 & 8.1e-3 & 2.1e-3 & 92.6 & 413.6 & 138.7 \\
 & LLaVA-1.5-7B & 0.595 & 0.405 & 7.2e-3 & 7.0e-4 & 117.3 & 576.0 & 67.7 \\
\cdashline{1-9}[1pt/5pt]
\multirow{2}{*}{Video} & Qwen2.5-VL-7B & 0.423 & 0.577 & 1.2e-2 & 2.0e-4 & 36.7 & 3584.0 & 86.4 \\
 & VideoLLaMA3-7B & 0.474 & 0.526 & 3.7e-2 & 2.6e-4 & 14.8 & 2283.0 & 15.8 \\
\cdashline{1-9}[1pt/5pt]
Audio & Qwen2-Audio-7B & 0.733 & 0.267 & 1.0e-2 & 3.2e-3 & 71.0 & 122.3 & 6.4 \\
\cdashline{1-9}[1pt/5pt]
Time Series & ChatTS-14B & 0.945 & 0.055 & 1.4e-2 & 3.2e-3 & 67.1 & 32.5 & 190.1 \\
\cdashline{1-9}[1pt/5pt]
Graph & GraphGPT-7B & 0.807 & 0.193 & 1.5e-3 & 1.0e-2 & 596.9 & 24.5 & 11.2 \\
\bottomrule
\end{tabular}
\end{table}

\begin{table}[ht]
\centering
\footnotesize
\setlength\tabcolsep{3pt}
\caption{Attention and token statistics for the \textit{last two layers}.}
\label{tab:consolidated_stats_last}
\begin{tabular}{ll|cc|cc|ccc}
\toprule
\multirow{2}{*}{\textbf{Modality}} & \multirow{2}{*}{\textbf{Model}} & \multicolumn{2}{c|}{\textbf{Total Attn}} & \multicolumn{2}{c|}{\textbf{Avg Attn/Tok}} & \multicolumn{3}{c}{\textbf{Avg Token Count}} \\
\cmidrule(lr){3-4} \cmidrule(lr){5-6} \cmidrule(lr){7-9}
 & & \textbf{Text} & \textbf{Non-t} & \textbf{Text} & \textbf{Non-t} & \textbf{Text} & \textbf{Non-t} & \textbf{Out} \\
\midrule
\multirow{3}{*}{Image} & Qwen2.5-VL-7B & 0.456 & 0.544 & 5.0e-3 & 1.5e-4 & 118.6 & 3584.0 & 208.0 \\
 & Kimi-VL-A3B & 0.474 & 0.526 & 7.0e-3 & 2.5e-3 & 92.6 & 413.6 & 138.7 \\
 & LLaVA-1.5-7B & 0.716 & 0.284 & 8.6e-3 & 4.9e-4 & 117.3 & 576.0 & 67.7 \\
\cdashline{1-9}[1pt/5pt]
\multirow{2}{*}{Video} & Qwen2.5-VL-7B & 0.448 & 0.553 & 1.3e-2 & 1.5e-4 & 33.8 & 3584.0 & 83.8 \\
 & VideoLLaMA3-7B & 0.491 & 0.509 & 3.7e-2 & 2.4e-4 & 15.2 & 2338.9 & 14.1 \\
\cdashline{1-9}[1pt/5pt]
Audio & Qwen2-Audio-7B & 0.507 & 0.493 & 7.1e-3 & 6.2e-3 & 71.0 & 80.0 & 20.0 \\
\cdashline{1-9}[1pt/5pt]
Time Series & ChatTS-14B & 0.924 & 0.076 & 1.4e-2 & 3.9e-3 & 67.1 & 32.5 & 191.0 \\
\cdashline{1-9}[1pt/5pt]
Graph & GraphGPT-7B & 0.860 & 0.140 & 1.6e-3 & 8.0e-3 & 596.9 & 24.5 & 11.2 \\
\bottomrule
\end{tabular}
\end{table}

\section{Case Study: Text Dominance in Action}
\label{app:case_study}
\begin{figure}[ht]
    \centering
    \includegraphics[width=\textwidth]{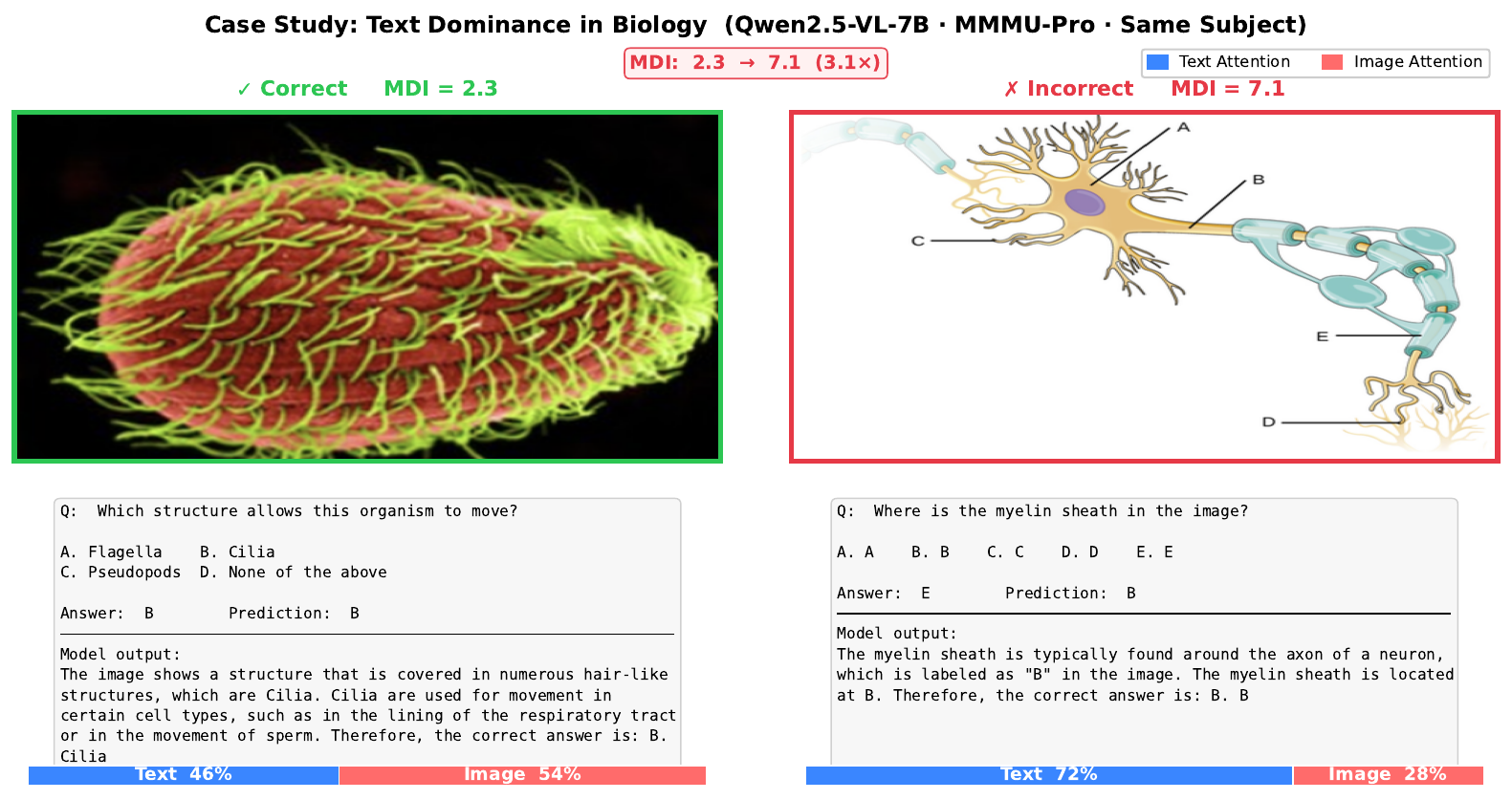}
    \caption{Case study of text dominance on two Biology questions (Qwen2.5-VL-7B, MMMU-Pro). \textbf{Left}: low MDI (2.3), the model attends to the image and answers correctly. \textbf{Right}: high MDI (7.1), the model over-relies on textual priors and misidentifies the labeled structure. The bottom bars show the text vs.\ image attention share.}
    \label{fig:case_study}
\end{figure}

To provide qualitative intuition for the quantitative findings in the main text, Figure~\ref{fig:case_study} contrasts two Biology questions from MMMU-Pro answered by Qwen2.5-VL-7B.

In Case~1 (MDI\,=\,2.3), the model allocates 54\% of its attention to image tokens, successfully identifies the hair-like cilia in the micrograph, and answers correctly.
In Case~2 (MDI\,=\,7.1, a 3.1$\times$ increase), 72\% of attention shifts to text tokens.
Faced with a labeled neuron diagram, the model falls back on textual prior knowledge (``the myelin sheath is typically found around the axon'') rather than reading the diagram labels, and incorrectly predicts B (axon) instead of the correct answer E.

Both questions belong to the same subject.
The contrast illustrates the mechanism suggested by our quantitative analysis: when text dominance is high, the model substitutes textual priors for visual evidence, leading to errors on questions that require genuine image understanding.

\section{Additional Token-Compression Results}
\label{app:extended_compression_revision}

Within each experiment, all conditions are evaluated on the same items. Scores for random dropping and random selection average three draws.

\subsection{Image Compression on MMBench-EN}
\label{app:high_baseline_revision}
We extend the image compression study to MMBench-EN~\citep{liu2024mmbench}, where LLaVA-1.5-7B achieves a higher accuracy with full visual tokens. The evaluation uses 500 held-out items, with a mean per-item chance accuracy of 27.38\%. Informed selection and random dropping use matched visual-token budgets.

\begin{table}[H]
\centering\small
\setlength{\tabcolsep}{4pt}
\caption{Image compression on 500 MMBench-EN items.}
\label{tab:high_baseline_revision}
\begin{tabular}{@{}lcc@{}}
\toprule
Condition & Removed & Accuracy (\%) \\
\midrule
Full visual tokens & 0\% & 73.4 \\
Informed selection & 50\% & 73.8 \\
Informed selection & 75\% & 73.6 \\
Random dropping & 75\% & 70.6 \\
Informed selection & 90\% & 72.6 \\
\bottomrule
\end{tabular}
\end{table}

Informed selection maintains accuracy close to the full-input result across the tested removal ratios. At 75\% removal, its advantage over random dropping demonstrates the role of token quality at a fixed token count.

\subsection{Video Compression on MVBench}
\label{app:video_compression_revision}
We evaluate Qwen2.5-VL-7B on 2,000 randomly sampled MVBench items~\citep{li2024mvbench}. At an intermediate LLM layer, informed selection retains the 50\% of video tokens receiving the most attention. Random selection uses the same token budget and pruning layer. The retention ratio is fixed on a separate development split before evaluation.

\begin{table}[H]
\centering\small
\setlength{\tabcolsep}{4pt}
\caption{Video compression on 2,000 randomly sampled MVBench items.}
\label{tab:video_compression_revision}
\begin{tabular}{@{}lccr@{}}
\toprule
Condition & Retained & Accuracy (\%) & MDI \\
\midrule
Full video & 100\% & 61 & 523.83 \\
Informed selection & 50\% & 64 & 271.68 \\
Random selection & 50\% & 58 & 313.70 \\
\bottomrule
\end{tabular}
\end{table}

At the same token budget and pruning layer, informed selection outperforms random selection. Compared with the full-video baseline, informed selection achieves higher accuracy and a lower late-layer MDI, indicating a smaller ratio of average attention per text token to that per video token.

\clearpage
\section*{NeurIPS Paper Checklist}

\begin{enumerate}

\item {\bf Claims}
    \item[] Question: Do the main claims made in the abstract and introduction accurately reflect the paper's contributions and scope?
    \item[] Answer: \answerYes{}
    \item[] Justification: The abstract and introduction state claims that are directly supported by the experimental results in Sections 3--5. We distinguish between controlled evidence and observational evidence throughout.

\item {\bf Limitations}
    \item[] Question: Does the paper discuss the limitations of the work performed by the authors?
    \item[] Answer: \answerYes{}
    \item[] Justification: A dedicated Limitations section discusses the non-causal nature of attention-based metrics, potential confounds in token replication, observational cross-model comparisons, and the scope of the compression experiments.

\item {\bf Theory assumptions and proofs}
    \item[] Question: For each theoretical result, does the paper provide the full set of assumptions and a complete (and correct) proof?
    \item[] Answer: \answerNA{}
    \item[] Justification: This paper does not include theoretical results or proofs. The contributions are empirical.

    \item {\bf Experimental result reproducibility}
    \item[] Question: Does the paper fully disclose all the information needed to reproduce the main experimental results of the paper to the extent that it affects the main claims and/or conclusions of the paper (regardless of whether the code and data are provided or not)?
    \item[] Answer: \answerYes{}
    \item[] Justification: All model names, benchmark names, metric definitions, compression ratios, and algorithmic details (Algorithm 1) are provided. All benchmarks and models used are publicly available.

\item {\bf Open access to data and code}
    \item[] Question: Does the paper provide open access to the data and code, with sufficient instructions to faithfully reproduce the main experimental results, as described in supplemental material?
\item[] Answer: \answerYes{}
    \item[] Justification: Code for attention extraction, metric computation, and token compression will be released upon acceptance. All datasets and models used are publicly available.
    \item[] Guidelines:
    \begin{itemize}
        \item The answer NA means that paper does not include experiments requiring code.
        \item Please see the NeurIPS code and data submission guidelines (\url{https://nips.cc/public/guides/CodeSubmissionPolicy}) for more details.
        \item While we encourage the release of code and data, we understand that this might not be possible, so ``No'' is an acceptable answer. Papers cannot be rejected simply for not including code, unless this is central to the contribution (e.g., for a new open-source benchmark).
        \item The instructions should contain the exact command and environment needed to run to reproduce the results. See the NeurIPS code and data submission guidelines (\url{https://nips.cc/public/guides/CodeSubmissionPolicy}) for more details.
        \item The authors should provide instructions on data access and preparation, including how to access the raw data, preprocessed data, intermediate data, and generated data, etc.
        \item The authors should provide scripts to reproduce all experimental results for the new proposed method and baselines. If only a subset of experiments are reproducible, they should state which ones are omitted from the script and why.
        \item At submission time, to preserve anonymity, the authors should release anonymized versions (if applicable).
        \item Providing as much information as possible in supplemental material (appended to the paper) is recommended, but including URLs to data and code is permitted.
    \end{itemize}

\item {\bf Experimental setting/details}
    \item[] Question: Does the paper specify all the training and test details (e.g., data splits, hyperparameters, how they were chosen, type of optimizer, etc.) necessary to understand the results?
    \item[] Answer: \answerYes{}
    \item[] Justification: Model names, benchmark splits, compression ratios, and layer groupings are specified in the main text. No training is performed; all experiments are inference-time evaluations.

\item {\bf Experiment statistical significance}
    \item[] Question: Does the paper report error bars suitably and correctly defined or other appropriate information about the statistical significance of the experiments?
    \item[] Answer: \answerNo{}
    \item[] Justification: We report accuracy and attention metrics without error bars. Results for random dropping and random selection are averaged over three draws.

\item {\bf Experiments compute resources}
    \item[] Question: For each experiment, does the paper provide sufficient information on the computer resources (type of compute workers, memory, time of execution) needed to reproduce the experiments?
    \item[] Answer: \answerYes{}
    \item[] Justification: Hardware specifications (NVIDIA A800 80\,GB GPU, Intel Xeon Gold 6348 CPU) and software versions are detailed in Appendix~\ref{app:env}. Inference latency is reported in Table~\ref{tab:reduction_compare_mdi_aei_acc_time}. All experiments are inference-only and can be reproduced on a single GPU.
    
\item {\bf Code of ethics}
    \item[] Question: Does the research conducted in the paper conform, in every respect, with the NeurIPS Code of Ethics \url{https://neurips.cc/public/EthicsGuidelines}?
    \item[] Answer: \answerYes{}
    \item[] Justification: This work analyzes publicly available models and benchmarks. It does not involve human subjects, personal data, or dual-use concerns.

\item {\bf Broader impacts}
    \item[] Question: Does the paper discuss both potential positive societal impacts and negative societal impacts of the work performed?
    \item[] Answer: \answerNA{}
    \item[] Justification: This work is a diagnostic study of attention allocation in existing models. It does not introduce new models or capabilities that pose direct societal risks.
    
\item {\bf Safeguards}
    \item[] Question: Does the paper describe safeguards that have been put in place for responsible release of data or models that have a high risk for misuse (e.g., pretrained language models, image generators, or scraped datasets)?
    \item[] Answer: \answerNA{}
    \item[] Justification: This paper does not release new models or datasets. The released code performs only diagnostic analysis on existing public models.

\item {\bf Licenses for existing assets}
    \item[] Question: Are the creators or original owners of assets (e.g., code, data, models), used in the paper, properly credited and are the license and terms of use explicitly mentioned and properly respected?
    \item[] Answer: \answerYes{}
    \item[] Justification: All models and benchmarks are cited with their original publications. All assets used are publicly available under permissive licenses.

\item {\bf New assets}
    \item[] Question: Are new assets introduced in the paper well documented and is the documentation provided alongside the assets?
    \item[] Answer: \answerNA{}
    \item[] Justification: This paper does not release new datasets or pretrained models. Evaluation code will be documented upon release.

\item {\bf Crowdsourcing and research with human subjects}
    \item[] Question: For crowdsourcing experiments and research with human subjects, does the paper include the full text of instructions given to participants and screenshots, if applicable, as well as details about compensation (if any)? 
    \item[] Answer: \answerNA{}
    \item[] Justification: This paper does not involve crowdsourcing or research with human subjects.

\item {\bf Institutional review board (IRB) approvals or equivalent for research with human subjects}
    \item[] Question: Does the paper describe potential risks incurred by study participants, whether such risks were disclosed to the subjects, and whether Institutional Review Board (IRB) approvals (or an equivalent approval/review based on the requirements of your country or institution) were obtained?
    \item[] Answer: \answerNA{}
    \item[] Justification: This paper does not involve human subjects research.

\item {\bf Declaration of LLM usage}
    \item[] Question: Does the paper describe the usage of LLMs if it is an important, original, or non-standard component of the core methods in this research? Note that if the LLM is used only for writing, editing, or formatting purposes and does not impact the core methodology, scientific rigorousness, or originality of the research, declaration is not required.
    \item[] Answer: \answerNA{}
    \item[] Justification: LLMs are the subject of study in this paper (we analyze their attention patterns), but they are not used as a non-standard component of our methodology.
    \item[] Guidelines:
    \begin{itemize}
        \item The answer NA means that the core method development in this research does not involve LLMs as any important, original, or non-standard components.
        \item Please refer to our LLM policy (\url{https://neurips.cc/Conferences/2026/LLM}) for what should or should not be described.
    \end{itemize}

\end{enumerate}

\end{document}